\documentclass[preprint,12pt]{elsarticle}

\usepackage[T1]{fontenc}
\usepackage[utf8]{inputenc}
\usepackage{lmodern,microtype}
\usepackage[a4paper,margin=22mm]{geometry}
\usepackage{amsmath,amssymb,mathtools,bm}
\usepackage{graphicx,booktabs,multirow}
\usepackage{algorithm,algpseudocode,placeins}
\usepackage[font=small,labelfont=bf]{caption}
\usepackage{xurl}
\usepackage[hidelinks]{hyperref}
\usepackage{needspace}
\hypersetup{pdftitle={HGTO: A Unified Graph-Based Physics-Informed Formulation for Structural Topology Optimization},
  pdfauthor={Kangzheng Liu; Uday Kumar Punna; Leixin Ma}}

\graphicspath{{figures/}}
\biboptions{sort&compress}
\journal{Knowledge-Based Systems}
\begin{document}

\begin{frontmatter}

\title{HGTO: A Unified Graph-Based Physics-Informed Formulation for Structural Topology Optimization}

\author[inst1]{Kangzheng Liu}
\author[inst1]{Uday Kumar Punna}
\author[inst1]{Leixin Ma\corref{cor1}}
\cortext[cor1]{Corresponding author.}

\affiliation[inst1]{
  organization={School for Engineering of Matter, Transport and Energy},
  addressline={Arizona State University},
  city={Tempe},
  postcode={85287},
  state={AZ},
  country={USA}}

\begin{abstract}
Density-based topology optimization is typically structured as a nested sequence of material updates, structural analyses, and sensitivity assessments. While neural density parameterization and dual-field physics-informed approaches provide data-free alternatives, most existing methods represent density and displacement as coordinate fields and make limited use of the discrete relationships inherent in the finite element mesh. The present study introduces HGTO, a unified graph-based formulation that extends complete neural topology optimization from coordinate space to finite-element graph space. Element densities are parameterized on the element graph derived from the mesh, and the structural state is determined on the corresponding node--element hypergraph. Finite element kinematics, numerical quadrature, constitutive response, and force assembly remain explicitly defined operations within the differentiable computation. The material field and equilibrium state are therefore coupled through a common finite-element incidence structure. Numerical studies show compliance comparable to conventional density-based optimization at substantially lower computational cost than a representative coordinate-based dual-field neural method. The same coupled formulation accommodates high-resolution and irregular meshes, three-dimensional structures, finite deformation, and elastoplastic response.
\end{abstract}

\begin{keyword}
Topology optimization \sep Graph neural network \sep Hypergraph neural network \sep Physics-informed learning \sep Finite element method \sep Neural parameterization
\end{keyword}

\end{frontmatter}

\section{Introduction}\label{sec:introduction}

Topology optimization (TO) seeks the most effective distribution of material within a prescribed design domain under given loads, boundary conditions, and design constraints. Since the pioneering work of Bends{\o}e and Kikuchi~\cite{bendsoe1988generating} and the subsequent development of the Solid Isotropic Material with Penalization (SIMP) method~\cite{bendsoe1989optimal,bendsoe2004topology}, density-based topology optimization has become one of the most established approaches for structural design. In a conventional SIMP formulation, the design domain is discretized into finite elements, an element-wise density field describes the material distribution, and finite element analysis is repeatedly performed to determine the structural response. The corresponding sensitivities are then used to update the design through the optimality criteria (OC) method, the Method of Moving Asymptotes (MMA), or related gradient-based algorithms~\cite{svanberg1987method}. Filtering and projection schemes are commonly introduced to suppress checkerboard artifacts, control feature sizes, and obtain nearly discrete material layouts~\cite{bourdin2001filters,guest2004achieving,wang2011projection}. Owing to decades of development in numerical algorithms and computational execution, SIMP continues to be a reliable and widely used method for both academic and engineering applications~\cite{sigmund2013topology,andreassen2011efficient}.

Nevertheless, conventional topology optimization is still a nested iterative procedure. Every modification of the density field changes the structural stiffness and therefore requires a new equilibrium analysis, followed by sensitivity evaluation and another design update. Repeated state solutions constitute a major part of the computational effort, particularly as the design resolution, number of loading cases, or physical complexity increases~\cite{aage2017giga,buhl2000stiffness}. In addition, design and mechanics are handled by different numerical procedures. The optimizer updates the density variables, a finite element solver evaluates the associated displacement field, and the resulting mechanical information is subsequently transferred back to the optimizer. Although this sequential organization is effective, it separates the representation of the material field from the calculation of the structural state, making it difficult to express the complete design problem within a single computational formulation.

The rapid development of deep learning has motivated extensive efforts to accelerate or reformulate topology optimization. A large class of methods uses data-driven neural networks to predict optimized material layouts directly from loads, boundary conditions, volume fractions, or partially converged designs. Convolutional neural networks, encoder--decoder models, generative adversarial networks, transfer-learning strategies, and diffusion models have demonstrated that near-optimal topologies can be generated in milliseconds or seconds once training has been completed~\cite{sosnovik2019neural,yu2019deep,nie2021topologygan,maze2023diffusion}. These methods are particularly attractive when many design problems are drawn from a fixed or closely related family of problems. Their performance, however, depends strongly on the availability and coverage of optimized training data. Generating the required datasets may itself require many conventional topology-optimization simulations, while unseen loading conditions, support configurations, geometries, or resolutions may markedly reduce prediction quality. Moreover, similarity to a reference topology at the image or density-field level does not necessarily ensure structural equilibrium, an effective load path, satisfaction of local constraints, or manufacturability.

To avoid dependence on databases of optimized structures, another class of methods uses neural networks as design parameterizations rather than predictors. In these approaches, the density field is represented as
\begin{equation}
    \rho(\mathbf{x})=f_{\boldsymbol{\theta}}(\mathbf{x}),
\end{equation}
where the network parameters $\boldsymbol{\theta}$ replace conventional element-wise design variables and are optimized for each problem using the structural objective and design constraints. Hoyer et al., TOuNN, TONR, and subsequent implicit-neural-representation methods demonstrated that a neural density field can be optimized without labeled topology data~\cite{hoyer2019neural,chandrasekhar2021tounn,zhang2021tonr}. Such representations provide shared design parameters, continuous evaluation of the density field, and control over its spatial frequency content. They can therefore be viewed as physics-driven neural reparameterizations of the design variables. Their optimization behavior, however, remains strongly influenced by the chosen network architecture and optimization scheme~\cite{woldseth2022artificial}. More importantly, these approaches alter only the design representation: the structural response is still obtained from a conventional finite element analysis, and the mechanics calculation remains external to the neural parameterization. Most neural density representations are also coordinate-based multilayer perceptrons and therefore infer spatial relationships from point coordinates without explicitly using the connectivity of the underlying finite element mesh.

Physics-informed neural networks have enabled a further transition from neural density parameterization towards complete neural topology optimization. PINNTO replaced the finite element state analysis with an energy-based neural solver rooted in the deep-energy formulation~\cite{samaniego2020energy,jeong2023pinnto}. NTopo represented both the density and displacement fields using implicit neural networks, while retaining sensitivity filtering and an OC-generated target density to stabilize the design update~\cite{zehnder2021ntopo}. CPINNTO combined a deep-energy displacement PINN with a neural density model and formulated the complete topology-optimization process without using labeled data or conventional finite element analysis~\cite{jeong2023complete}. DMF-TONN further demonstrated that the density and displacement networks could be connected more directly, allowing the density representation to be updated without sensitivity filtering, OC updates, or repeated fitting to target densities~\cite{joglekar2024dmf}. More recently, DPNN-TO adopted a variational dual-network formulation with a sinusoidal displacement network and a Fourier-enhanced density network, extending the approach to high-resolution two- and three-dimensional problems, multiple loads, and multiple design constraints~\cite{singh2026dpnn}. These developments established an important new interpretation of topology optimization: the material field and the structural state can be treated as coupled unknowns and determined within the same physics-driven optimization process.

The principal difficulty in these complete and dual-field formulations lies in approximating the structural state. Density and displacement are generally represented by coordinate-based neural networks and coupled via an energy formulation, a residual loss, or a design objective. Whenever the density field changes, the displacement network must be retrained or further optimized to recover the corresponding equilibrium solution. The accuracy of the topology update, therefore, depends directly on the quality of the neural state approximation. Errors in displacement, strain, stress, or numerical integration carry over to the compliance and design gradient. This dependence becomes more pronounced near solid--void interfaces, stress concentrations, complex boundaries, low-volume-fraction designs, and nonlinear constitutive regimes. Present methods consequently rely on problem-dependent choices of sampling points, network size, frequency range, learning rate, loss weights, and the number of displacement-training iterations. Inaccurate state solutions may also produce asymmetric or mechanically inconsistent designs in otherwise symmetric problems~\cite{jeong2023complete,joglekar2024dmf,singh2026dpnn}.

A more essential limitation is that coordinate-based dual-field methods make limited use of the discrete mechanical structure already available in the computational mesh. The displacement network receives spatial coordinates and must recover displacement gradients, local material response, and equilibrium from sampled points. Element adjacency, node--element incidence, numerical quadrature, constitutive variables, and the assembly of element forces are not inherent parts of the neural representation. Existing complete PINN methods, therefore, integrate design and mechanics at the level of the optimization objective but not at the level of the underlying finite element structure. Their mesh-free representation provides flexibility in spatial sampling but also discards connectivity information, which is particularly useful for irregular domains, unstructured meshes, and element-based constitutive calculations.

Graph neural networks (GNNs) provide a natural means of retaining this information. A finite element mesh is inherently relational: neighboring elements define the local neighborhood of the material field, nodes carry mechanical degrees of freedom, and each finite element establishes a multi-node mechanical relation. Recent GNN-based topology-optimization methods have exploited this structure for near-optimal topology prediction over irregular domains, convergence speedup on unstructured meshes, as well as graph-based density regularization~\cite{seo2023graph,joo2024dynamic,gavris2024topology}. More recent graph neural-field methods directly parameterize element densities on the finite-element mesh and optimize the GNN using structural objectives derived from differentiable finite-element analysis~\cite{tabarraei2026differentiable,bhuiyan2026support}. These studies show that mesh connectivity provides an effective basis for parameterizing neural density. Nevertheless, the graph is used primarily to represent the material field, whereas the equilibrium problem is still solved by a separate matrix-based finite element procedure.

In parallel, graph-based computational mechanics has increasingly incorporated the discrete structure of numerical mechanics. Physics-informed graph neural Galerkin networks introduced graph-based variational formulations for irregular shapes and unstructured meshes~\cite{gao2022graphgalerkin}. Graph-convolutional deep-energy methods additionally demonstrated that finite-element shape-function gradients can improve robustness relative to purely coordinate-based automatic differentiation in demanding deformation problems~\cite{he2023graphdem}. Finite-element-inspired hypergraph networks then introduced an explicit node--element representation to preserve the higher-order relations of finite element meshes~\cite{gao2024finite}. Most recently, the FEM-Informed Hypergraph Neural Network (FHGNN) formulated isoparametric mapping, shape-function gradients, Gauss-point strain and stress evaluation, constitutive updates, numerical integration, and internal-force assembly as prescribed node-to-element and element-to-node operations~\cite{yang2026fhgnn}. Unlike a data-driven surrogate, these operations are determined by the finite element formulation rather than learned from simulation data. FHGNN therefore provides a differentiable finite-element-consistent representation of structural mechanics and has been applied to three-dimensional and path-dependent elastoplastic problems. Its current formulation, however, addresses forward structural analysis with a prescribed material distribution.

The graph representations used for material design and finite element mechanics are closely related. The element graph used to propagate density information is the dual representation induced by the node--element incidence of the same finite element mesh. The same incidence relation also transfers nodal displacements to an element, evaluates the corresponding element response, and assembles the element forces back to the nodes. Thus, the graph used for density parameterization and the hypergraph used for structural mechanics are complementary views of the same finite element discretization.

Taken together, these developments point to a more fundamental opportunity for physics-informed topology optimization: replacing the coordinate-based coupling of design and state with a mesh-native formulation derived directly from the finite element discretization. In existing dual-field approaches, density and displacement are coupled through the optimization objective, but the geometric and mechanical relations carried by the computational mesh remain largely outside the neural representation. Graph-based formulations, on the other hand, have shown that these relations can be retained explicitly and used as the computational structure for both material representation and structural mechanics. This suggests that the finite element mesh itself can serve as a common basis for describing material evolution and equilibrium, allowing the two fields to interact through the same discrete mechanical structure rather than through independently parameterized coordinate spaces. To the best of the authors’ knowledge, such a mesh-native formulation of complete physics-informed topology optimization has not yet been established.

To address this gap, we introduce HGTO, a unified graph-based physics-informed formulation for structural topology optimization. HGTO advances complete neural topology optimization by transitioning from coordinate-based field representations to finite-element graphs. The density field is parameterized on the element graph derived from the mesh, and the structural state is evaluated on the corresponding node--element hypergraph using prescribed finite-element operations. Since both representations are based on the same finite-element incidence structure, material evolution and mechanical equilibrium are integrated within a single mesh-connected differentiable formulation. Consequently, HGTO maintains the data-free optimization advantages of neural design parameterization while explicitly preserving mesh connectivity and the local mechanics inherent to the finite element discretization.

The main contributions of this work are summarized as follows:
\begin{enumerate}
    \item A complete graph-based formulation of physics-informed topology optimization is introduced, extending the coupled density--displacement paradigm from coordinate neural fields to finite element graph representations derived from a common mesh incidence structure.

    \item Finite element mechanics is retained explicitly within the differentiable graph computation. Material interpolation, finite-element kinematics, numerical quadrature, constitutive evaluation, and force assembly are prescribed operations within the same optimization process used to update the neural density representation.

    \item The common mesh-based representation preserves element connectivity and node--element relations throughout the optimization, providing a consistent basis for irregular design domains, unstructured discretizations, and different constitutive descriptions without changing the overall formulation.
\end{enumerate}

The proposed framework is evaluated using a series of structural topology-optimization problems, with comparisons to conventional density-based topology optimization and the coordinate-based dual-field neural approach. The numerical studies examine the accuracy of mechanical solutions, the quality and convergence of optimized topologies, and the computational characteristics of the coupled design and analysis process.

The remainder of this paper is organized as follows. Section~\ref{sec:hgto} presents the HGTO formulation. Sections~\ref{sec:experiments} and \ref{sec:nonlinear} examine linear and nonlinear design problems, respectively. Section~\ref{sec:discussion} discusses the findings and concludes with the limitations and future directions of the framework.

\section{HGTO: coupled topology and physics fields}
\label{sec:hgto}

HGTO describes material distribution and structural equilibrium on two graph fields derived from the same finite-element mesh. The topology-field graph PINN generates element densities through learnable message passing. The physics-field hypergraph PINN uses these densities in its constitutive response and updates its nodal states to satisfy equilibrium. Element sensitivities then connect the mechanical response to the parameters of the topology field.

\subsection{Problem formulation}
\label{subsec:problem}
Consider a design domain $\Omega\subset\mathbb R^d$, with $d=2$ or 3, discretized into $N_e$ elements and $N_n$ nodes. For a prescribed force vector $\bm f$ and homogeneous displacement supports, the minimum-compliance problem is
\begin{equation}
\begin{aligned}
\min_{\bm\rho}\quad &C(\bm\rho)=\bm f^{\mathsf T}\bm u^*,\\
\text{subject to}\quad &\bm K(\bm\rho)\bm u^*=\bm f,\\
&\sum_{e=1}^{N_e}V_e\rho_e\leq V_fV_\Omega,\qquad
\rho_{\min}\leq\rho_e\leq1,
\end{aligned}
\label{eq:problem}
\end{equation}
where $V_e$ is the volume of element $e$, $V_\Omega=\sum_eV_e$, and $V_f$ is the prescribed material fraction. Displacements and equilibrium equations are expressed on the free degrees of freedom. The SIMP interpolation defines the element modulus and global stiffness as
\begin{equation}
E_e(\rho_e)=E_{\min}+(E_0-E_{\min})\rho_e^p,
\qquad
\bm K(\bm\rho)=\sum_e\bm P_e^{\mathsf T}
\big[E_e(\rho_e)\widehat{\bm k}_e\big]\bm P_e.
\label{eq:simp}
\end{equation}
Here $E_0$ and $E_{\min}>0$ are the solid modulus and stiffness floor, $p$ is the penalization exponent, $\widehat{\bm k}_e$ is the element stiffness at unit modulus, and $\bm P_e$ gathers the local displacement $\bm u_e=\bm P_e\bm u$. The available material is fully used in these linear compliance problems. HGTO therefore imposes the volume equality through the density parameterization introduced below. Section~\ref{sec:nonlinear} extends the state equation to finite-deformation and elastoplastic response.

\subsection{Dual-field representation of the mesh}
\label{subsec:dual_graph}
\begin{figure}[!t]
\centering\includegraphics[width=.97\linewidth]{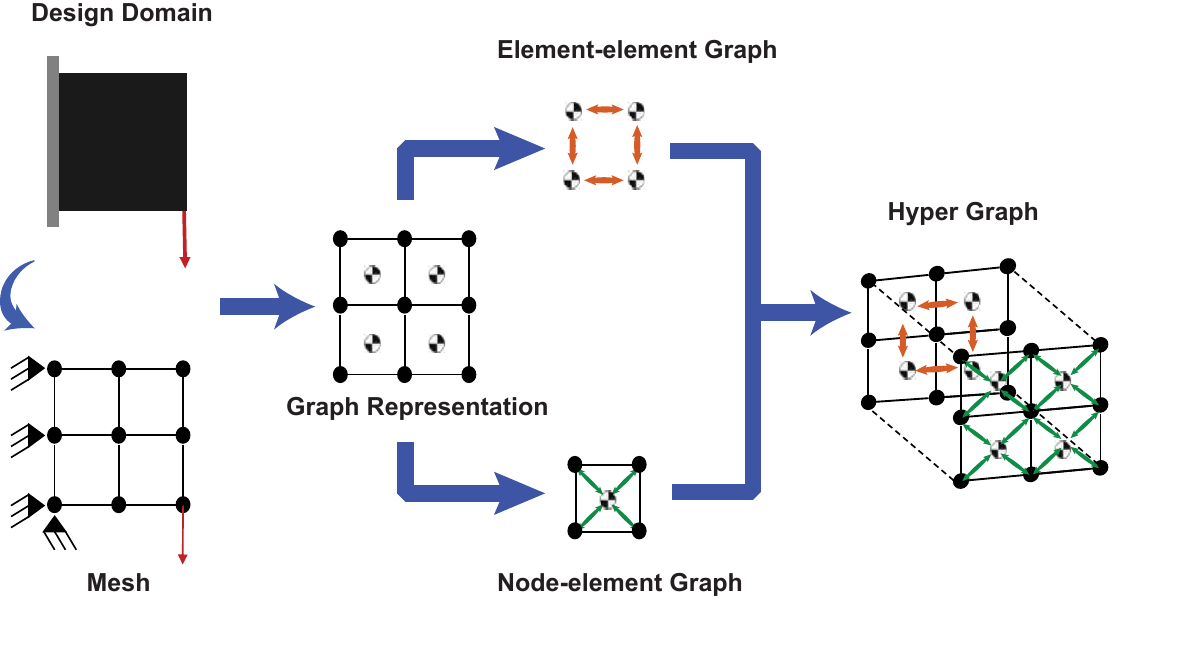}
\caption{Graph representations of the same finite-element mesh. An element is a node in the topology graph $\mathcal G_\rho$ and a hyperedge in the physics hypergraph $\mathcal G_u$.}
\label{fig:dual_field_graph}
\end{figure}

The incidence matrix $\bm H\in\{0,1\}^{N_n\times N_e}$ records the membership of mesh nodes in finite elements: $H_{ie}=1$ when node $i$ belongs to element $e$. Together with the ordered local connectivity, it defines the complementary graph views in Fig.~\ref{fig:dual_field_graph}. In the topology graph $\mathcal G_\rho=(\mathcal V_e,\mathcal E_e)$, each element is a graph node. Adjacent elements share a complete edge in two dimensions or a complete face in three dimensions. Their centroids provide spatial features, and their connections determine the neighborhoods used to update these features.

The physics hypergraph $\mathcal G_u=(\mathcal V_n,\mathcal E_h,\bm H)$ associates displacement states with mesh nodes and represents each element as a hyperedge joining its constituent nodes. Density and constitutive parameters are stored with the element, together with its geometry and quadrature data. This representation retains the multi-node interactions of finite-element mechanics. An element density generated on $\mathcal G_\rho$ is available at the corresponding hyperedge of $\mathcal G_u$, and its mechanical sensitivity returns through the same index.

\begin{figure}[!t]
\centering\includegraphics[width=\linewidth]{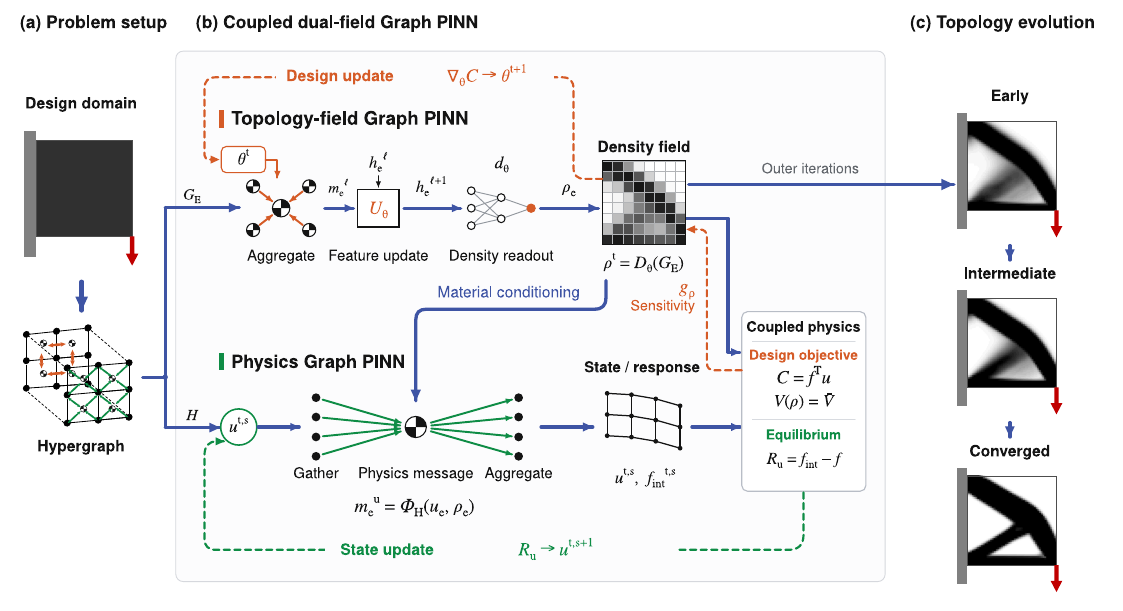}
\caption{HGTO workflow. The topology graph generates physical densities, and the physics hypergraph evaluates the structural response. Element sensitivities are propagated back to update the topology network. The right-hand panels show the evolving design.}
\label{fig:hgto_workflow}
\end{figure}

Figure~\ref{fig:hgto_workflow} presents the optimization loop. The topology field generates the current physical density, the physics field determines the associated equilibrium state, and the design gradient updates the topology-network parameters. Connectivity and reference geometry remain fixed as the material field and mechanical states evolve on the common discretization.

\subsection{Topology-field graph PINN}
\label{subsec:topology_field}
\subsubsection{Graph parameterization}
The topology network assigns one scalar output to each element. Normalized centroids $\bm c_e\in[0,1]^d$ are embedded using the Fourier features
\begin{equation}
\bm h_e^{(0)}=\bm\gamma(\bm c_e)=
\begin{bmatrix}\sin(2\pi\bm F\bm c_e)\\\cos(2\pi\bm F\bm c_e)\end{bmatrix},
\qquad F_{ab}\sim\mathcal N(0,\sigma_F^2),
\label{eq:fourier_features}
\end{equation}
where $\bm F\in\mathbb R^{n_F\times d}$ is sampled at initialization and then held fixed. Two first-degree Chebyshev graph convolutions~\cite{defferrard2016convolutional} propagate the features over adjacent elements:
\begin{equation}
\begin{aligned}
\bm h^{(1)}&=\operatorname{ReLU}\!\left(
\bm h^{(0)}\bm W_0^{(0)}+\widetilde{\bm L}\bm h^{(0)}\bm W_1^{(0)}+\bm b^{(0)}\right),\\
\bm z&=\bm h^{(1)}\bm W_0^{(1)}+
\widetilde{\bm L}\bm h^{(1)}\bm W_1^{(1)}+\bm b^{(1)}.
\end{aligned}
\label{eq:topology_network}
\end{equation}
The normalized graph Laplacian $\bm L$ is rescaled as $\widetilde{\bm L}=2\bm L/\lambda_{\max}-\bm I$, using the spectral bound $\lambda_{\max}=2$. Each layer combines the features of an element with those of its neighbors, and the second layer produces the scalar logits $z_e$. The trainable parameters $\bm\theta$ comprise the shared weights and biases. The parameter count is independent of mesh resolution: for hidden width $h$, the two layers contain $N_\theta=4n_Fh+3h+1$ trainable parameters. The fixed Fourier features and their first-layer graph propagation can be cached for a given mesh.

\subsubsection{Density filtering and volume constraint}
The network outputs are converted to physical densities by spatial averaging and smooth projection. The filter averages nearby design values within a radius $r_f$, with weights determined by distance and element volume. For the logits $\bm z$, the density map is
\begin{equation}
\begin{aligned}
\bm x&=\operatorname{sigmoid}(\bm z-\lambda_V\bm1),\\
\bm\rho&=\rho_{\min}\bm1+(1-\rho_{\min})\mathcal H_\beta(\bm A\bm x),
\end{aligned}
\label{eq:density_projection}
\end{equation}
where $\bm A$ is the normalized filter matrix and $\mathcal H_\beta$ is a smooth Heaviside projection. Increasing $\beta$ sharpens the solid--void boundary. The common shift $\lambda_V$ is chosen to satisfy the material constraint after filtering and projection:
\begin{equation}
\sum_e V_e\rho_e(\bm z,\lambda_V)=V_fV_\Omega.
\label{eq:volume_root}
\end{equation}
Gradients are propagated through the filter, projection, and volume constraint when updating the topology network. The filter weights, projection function, and volume-constrained derivative are given in Section S1 of the supplementary material.

\subsection{Hypergraph physics-field PINN}
\label{subsec:physics_field}
The physics field stores displacement as nodal attributes. For a fixed density field, a forward pass gathers the nodal attributes at each hyperedge, evaluates the element response, and returns force contributions to the nodes:
\begin{equation}
\bm u_e=\bm P_e\bm u,\qquad
\bm m_e^u=\Phi_e(\bm u_e,\rho_e),\qquad
\bm f^{\mathrm{int}}=\sum_e\bm P_e^{\mathsf T}\bm m_e^u.
\label{eq:physics_messages}
\end{equation}
The physical message map $\Phi_e$ contains the finite-element kinematics, constitutive response, and numerical integration. For small-strain elasticity,
\begin{equation}
\bm\varepsilon_{eg}=\bm B_{eg}\bm u_e,\qquad
\bm\sigma_{eg}=\bm D_e(\rho_e)\bm\varepsilon_{eg},\qquad
\bm m_e^u=\sum_g\bm B_{eg}^{\mathsf T}\bm\sigma_{eg}\,\omega_{eg},
\label{eq:element_mechanics}
\end{equation}
where $\bm B_{eg}$ is the strain--displacement matrix at quadrature point $g$ and $\omega_{eg}$ includes the quadrature weight and Jacobian determinant. The resulting message is $\bm m_e^u=E_e\widehat{\bm k}_e\bm u_e$, and its associated strain energy is $\mathcal W_e=\tfrac12\bm u_e^{\mathsf T}\bm m_e^u$.

Structural equilibrium is obtained by minimizing the total potential energy,
\begin{equation}
\begin{aligned}
\mathcal L_u(\bm u;\bm\rho)&=\sum_e\mathcal W_e-\bm f^{\mathsf T}\bm u,\\
\nabla_{\bm u}\mathcal L_u&=\bm f^{\mathrm{int}}-\bm f=\bm R,\qquad
\bm u^*(\bm\rho)=\arg\min_{\bm u}\mathcal L_u.
\end{aligned}
\label{eq:physics_loss}
\end{equation}
The free nodal displacements are the optimization variables. The message operators are prescribed by the element formulation, and displacement supports are imposed directly. Element geometry, quadrature data, and the incidence maps are cached before optimization.

For linear elasticity, sufficient displacement constraints and a positive stiffness floor make $\mathcal L_u$ a strictly convex quadratic. Preconditioned conjugate gradients minimize this loss using node--element--node operator actions, with geometric or algebraic multigrid adapted to the mesh structure. For the nonlinear cases in Section~\ref{sec:nonlinear}, equilibrium is followed incrementally using the finite-deformation and elastoplastic operators derived in Sections S3 and S4 of the supplementary material.

\subsection{Coupling and optimization}
\label{subsec:coupling}
The topology field is trained using the equilibrium compliance,
\begin{equation}
\mathcal L_\rho(\bm\theta)=
\frac{\bm f^{\mathsf T}\bm u^*(\bm\rho(\bm\theta))}{C_{\mathrm{ref}}},
\label{eq:design_loss}
\end{equation}
where $C_{\mathrm{ref}}$ is the initial compliance, held fixed during optimization. The volume constraint is incorporated through Eq.~\eqref{eq:volume_root}. For the fixed-load linear problem, differentiating equilibrium gives the element sensitivity and the network gradient:
\begin{equation}
\begin{aligned}
g_{\rho,e}=\frac{dC}{d\rho_e}
&=-p\rho_e^{p-1}(E_0-E_{\min})
(\bm u_e^*)^{\mathsf T}\widehat{\bm k}_e\bm u_e^*,\\
\nabla_{\bm\theta}\mathcal L_\rho
&=\frac{1}{C_{\mathrm{ref}}}
\sum_e g_{\rho,e}\nabla_{\bm\theta}\rho_e.
\end{aligned}
\label{eq:design_gradient}
\end{equation}
Each sensitivity is evaluated at its corresponding physics hyperedge and passed to the aligned element output of the topology graph. Automatic differentiation propagates these sensitivities through the constrained density map and graph layers.

Continuation in the material penalization and projection sharpness guides the density field from a diffuse initial distribution toward the final design. Adam updates the network parameters using the equilibrium objective and its gradient through a sequence of prescribed $(p,\beta)$ stages. Once the terminal values are reached, optimization continues until both the objective and physical density stabilize. The nonlinear examples use the same density--state coupling with incremental mechanical feedback and the schedules specified in Section~\ref{sec:nonlinear}.

Algorithm~\ref{alg:hgto} summarizes the procedure. Convergence is assessed on volume-feasible, equilibrated designs at the final material parameters. A maximum update count limits the computation. The final design is evaluated with the terminal material parameters and state tolerance.

\begin{algorithm}[!htbp]
\caption{HGTO optimization}
\label{alg:hgto}
\small
\begin{algorithmic}[1]
\Require Mesh, supports, loading, material model, $V_f$, and optimization schedule
\Ensure Physical density $\bm\rho^*$ and equilibrium state
\State Construct $\mathcal G_\rho$, $\mathcal G_u$, and filter $\bm A$; cache element data
\State Initialize the topology parameters $\bm\theta$ and mechanical state
\For{each design update within the prescribed maximum}
  \State Select $p$ and $\beta$ from the continuation schedule
  \State Evaluate logits $\bm z$ and solve Eq.~\eqref{eq:volume_root} for the physical density $\bm\rho$
  \State Solve equilibrium, including the loading history when required
  \State Evaluate the objective and physical-density sensitivities $\bm g_\rho$
  \If{this is the initial evaluation}\State Set $C_{\mathrm{ref}}\gets C$\EndIf
  \State Backpropagate $\bm g_\rho/C_{\mathrm{ref}}$ through the density map and graph network
  \State Update $\bm\theta$ with Adam using the objective gradient
  \If{the final-stage objective and density meet the stopping tolerances}\State \textbf{break}\EndIf
\EndFor
\State Evaluate the final density at the prescribed terminal material parameters and state tolerance
\State \Return $\bm\rho^*$ and its equilibrium state
\end{algorithmic}
\end{algorithm}

\FloatBarrier
\section{Results}
\label{sec:experiments}
\label{sec:linear}
The numerical studies examine the design quality, computational cost, and applicability of HGTO in comparison with conventional density-based optimization and coordinate-based dual-field neural TO. We begin with planar benchmarks to assess the quality and cost of the optimized designs, then examine finer discretizations, irregular domains, and three-dimensional structures. Finite-deformation and elastoplastic problems are considered in Section~\ref{sec:nonlinear}.

\subsection{Experimental settings}
\label{subsec:settings}
SIMP--OC serves as the conventional reference, and NTopo~\cite{zehnder2021ntopo} represents coordinate-based dual-field neural TO. NTopo follows the original formulation and training settings, with case-specific settings summarized in the supplementary material.

The design domain, loading, supports, and prescribed material fraction are held fixed within each numerical comparison. Final linear designs are evaluated with a common finite-element model implemented in scikit-fem~\cite{GustafssonMcBain2020ScikitFEM}. The linear material parameters are $E_0=1$, $E_{\min}=10^{-6}$, and $\nu=0.3$. HGTO and SIMP--OC use the same density filter and projection.

The neural computations run on an NVIDIA RTX 6000 Ada GPU. SIMP--OC uses two CPU threads on an AMD Ryzen Threadripper PRO 5995WX. Network configurations, stopping criteria, achieved volumes, and solver settings are given in the supplementary material. Each neural design is optimized for its prescribed problem without labeled topology data.

\subsection{Two-dimensional benchmarks}
\label{subsec:benchmark2d}
\begin{figure}[!htbp]
\centering\includegraphics[width=\linewidth]{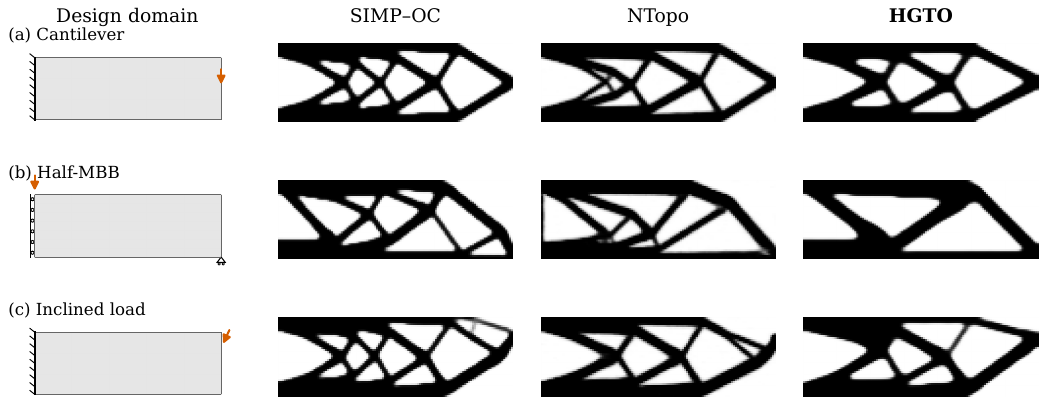}
\caption{Optimized topologies for (a) a centrally loaded cantilever, (b) a half-MBB beam, and (c) a cantilever under inclined loading. The prescribed volume fraction is 0.50.}
\label{fig:benchmark2d}
\end{figure}

The first examples are a centrally loaded cantilever, a half-MBB beam, and a cantilever subjected to an inclined eccentric load. Each uses a $120\times40$ mesh and a material fraction of 0.50. Figure~\ref{fig:benchmark2d} shows that HGTO recovers the principal members connecting the loads and supports, with compliance comparable to the conventional and dual-field neural references (Table~\ref{tab:linear}).

The designs differ more in the subdivision of their interiors than in their stiffness. For the half-MBB beam, HGTO places material in a few broad diagonal members and leaves larger open regions. The two reference methods introduce additional diagonals and smaller internal cells. The inclined-load HGTO design similarly contains fewer slender secondary members, particularly toward the loaded end. These arrangements retain the main load-bearing connections while concentrating material in fewer members. The centrally loaded cantilever shows closer agreement among all three methods, including the upper and lower chords and their diagonal connections.

The cost of obtaining these designs differs substantially. HGTO completes the three optimizations in 7.4--8.5~s, whereas NTopo requires 862--943~s. The graph-based formulation thus retains a learned material representation at a total cost of the same order as SIMP--OC on these small problems. The shared mesh representation connects the learned density field directly to the nodal equilibrium states used in the design update.

\begin{table}[!htbp]
\centering\small
\caption{Compliance $C$ and total optimization time $t$ (s) for the linear examples. $\dagger$ Update limit reached.}
\label{tab:linear}
\setlength{\tabcolsep}{5pt}
\begin{tabular}{lrrrrrr}
\toprule
 & \multicolumn{2}{c}{SIMP--OC} & \multicolumn{2}{c}{NTopo} & \multicolumn{2}{c}{HGTO} \\
\cmidrule(lr){2-3}\cmidrule(lr){4-5}\cmidrule(lr){6-7}
Case & $C$ & $t$ & $C$ & $t$ & $C$ & $t$ \\
\midrule
Cantilever & 175.08 & 9.0 & 177.93 & 862.2 & 176.49 & 8.3 \\
Half-MBB & 191.49 & 9.0 & 194.79 & 942.5 & 195.47 & 7.4 \\
Inclined load & 132.72 & 26.9$^{\dagger}$ & 133.81 & 906.6 & 133.48 & 8.5 \\
\addlinespace[3pt]
L-bracket & 96.06 & 9.7 & 99.43 & 952.3 & 96.95 & 10.3 \\
Perforated bracket & 49.32 & 6.3 & 49.78 & 1001.1 & 48.71 & 14.2 \\
\addlinespace[3pt]
3D cantilever & 15.07 & 188.3 & 16.17 & 3886.0 & 15.23 & 24.3 \\
Four-foot support & 0.5374 & 377.2 & 0.5810 & 3781.3 & 0.5347 & 39.3 \\
Torsion ($C\times10^3$) & 8.135 & 337.0 & 10.148 & 3564.1 & 8.144 & 48.1 \\
\bottomrule
\end{tabular}
\end{table}

\subsection{High-resolution optimization}
\label{subsec:resolution}
HGTO and SIMP--OC optimize the cantilever on meshes ranging from $120\times40$ to $960\times320$ elements. HGTO uses the same network configuration and 16,577 trainable parameters at every resolution. Refinement increases the number of density outputs and mechanical states, while the graph layers continue to share their weights over the mesh.

\begin{figure}[!htbp]
\centering\includegraphics[width=\linewidth]{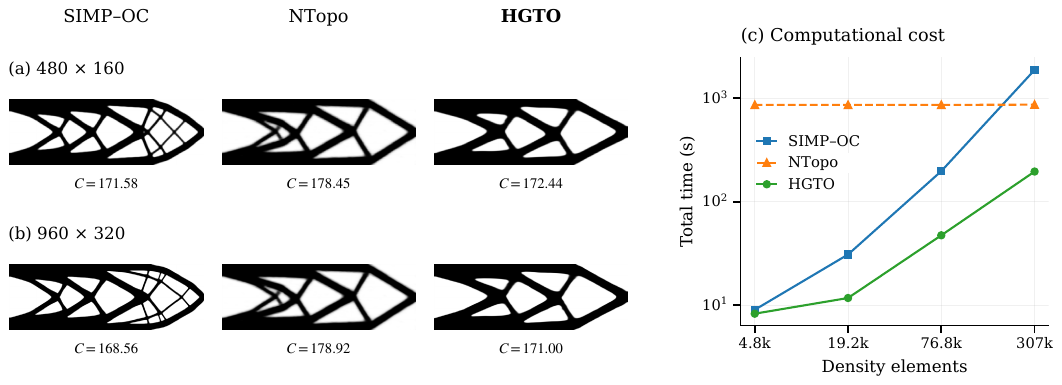}
\caption{Cantilever designs on the (a) $480\times160$ and (b) $960\times320$ meshes, and (c) total computation time.}
\label{fig:resolution}
\end{figure}

The fine-resolution designs in Fig.~\ref{fig:resolution} have similar compliance but different internal layouts. SIMP--OC develops a network of thin branches near the loaded end. HGTO retains fewer principal diagonals, with well-defined boundaries and junctions. Increasing the resolution resolves these members more closely without requiring a more heavily subdivided interior. The compact parameterization thus accommodates a detailed density field while preserving the broad organization of the material.

The runtime separation grows with problem size. HGTO takes 195.7~s on the finest mesh, containing 307,200 density elements, compared with 1,880.1~s for the CPU SIMP--OC implementation. The speedup is approximately fourfold at $480\times160$ and nearly tenfold at $960\times320$.

\subsection{Irregular design domains}
\label{subsec:irregular}
An L-bracket and a perforated bracket are considered at a material fraction of 0.40. The former introduces a re-entrant boundary, while the latter has an internal circular clearance and a nonuniform mesh. The topology-network architecture is retained, with its neighborhoods constructed from the element connections in each domain.

\begin{figure}[!htbp]
\centering\includegraphics[width=\linewidth]{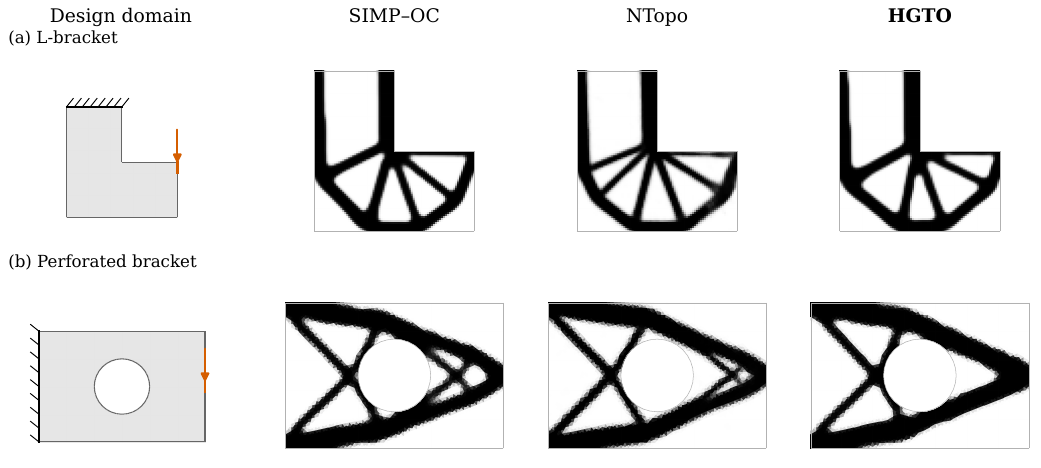}
\caption{Optimized topologies for (a) the L-bracket and (b) the perforated bracket at a prescribed volume fraction of 0.40.}
\label{fig:irregular}
\end{figure}

HGTO and SIMP--OC produce closely matched compliance in both cases. In the L-bracket, the vertical members join a fan of inclined members around the re-entrant region (Fig.~\ref{fig:irregular}a). This arrangement is similar across the three methods. In the perforated bracket, material separates into upper and lower paths around the clearance and rejoins near the load. HGTO leaves a larger open region between the hole and the loaded end, whereas the reference layouts introduce smaller internal branches there. The simpler subdivision preserves the overall stiffness and the required clearance.

The change in geometry is handled through the mesh relations already used by the two fields. The topology graph follows the element neighborhoods around the opening, and the physics hypergraph retains the corresponding nodal connections. This correspondence also accommodates unequal element sizes in the perforated domain. HGTO obtains the two designs in 10--14~s, compared with approximately 16~min per case for the dual-field neural baseline. The common mesh representation accommodates these geometric changes at a cost of the same order as SIMP--OC.

\subsection{Three-dimensional structures}
\label{subsec:three_dimensional}
The three-dimensional examples comprise an end-loaded cantilever, a four-foot support, and a torsion member. The three cases use a common 3D network configuration and require different spatial arrangements of material. HGTO achieves compliance comparable to SIMP--OC throughout this set (Table~\ref{tab:linear}).

\begin{figure}[!htbp]
\centering\includegraphics[width=.97\linewidth]{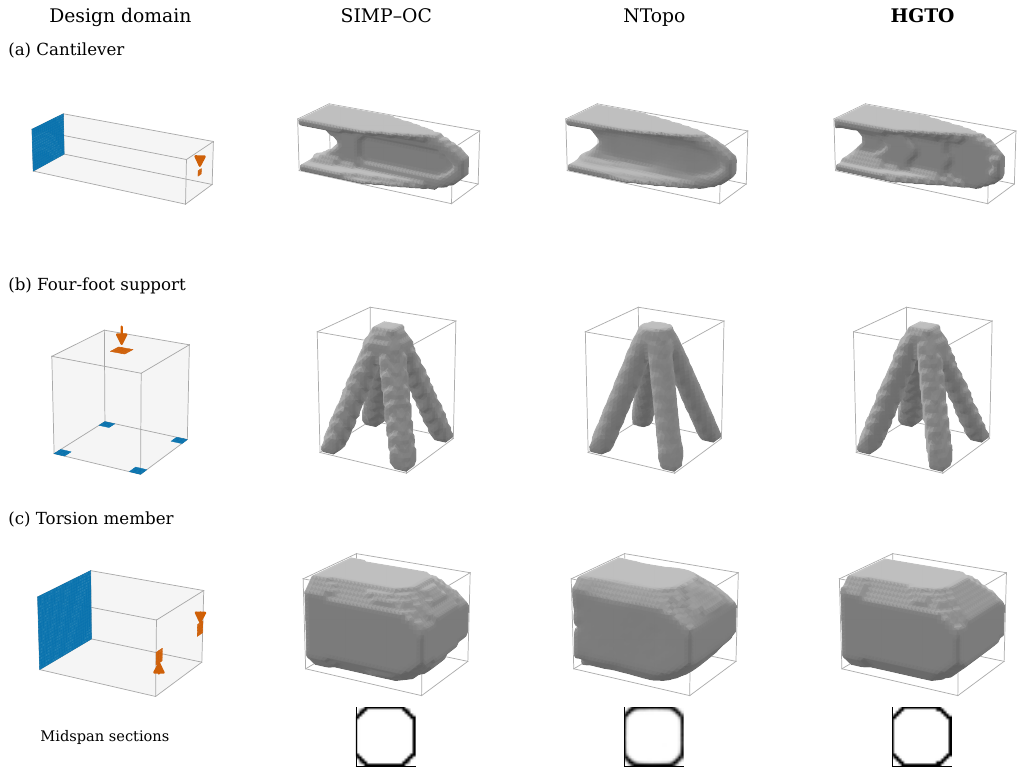}
\caption{Three-dimensional topologies for (a) the cantilever, (b) the four-foot support, and (c) the torsion member. Surfaces are shown at $\rho=0.5$; the insets give midspan density sections of the torsion designs.}
\label{fig:three_dimensional}
\end{figure}

Figure~\ref{fig:three_dimensional} shows how these arrangements change with the loading. The cantilever places material in upper and lower longitudinal members linked by webs. The four-foot support divides the load from the upper pad among four inclined legs. The torsion member places material around a hollow interior, forming the closed section visible in the midspan insets. The learned density representation accommodates both branching members and continuous walls, with the resulting forms closely following those obtained by SIMP--OC.

NTopo captures the broad exterior shapes, although its torsion design is about 25\% more compliant than HGTO in the common evaluation. The section views show a more diffuse material boundary for NTopo, while HGTO and SIMP--OC have more sharply defined walls. For this case, the distribution of material through the section distinguishes the designs more clearly than their similar exterior envelopes.

HGTO completes the 3D optimizations in 24--48~s. SIMP--OC takes 188--377~s, and NTopo approximately one hour. Relative to this dual-field neural baseline, HGTO reduces runtime by factors of 74--160. The efficiency of the coupled graph formulation therefore extends to volumetric density fields, where both the structural state and the material distribution contain substantially more variables.

\FloatBarrier
\section{Nonlinear topology optimization}
\label{sec:nonlinear}
The following examples extend the coupled graph formulation to finite deformation and elastoplasticity. The material field is generated by the same topology-network construction, while the physical operators on the hypergraph account for the nonlinear response. The constitutive models, incremental state equations, and density sensitivities are derived in Sections S3 and S4 of the supplementary material.

\subsection{Finite deformation under increasing load}
A cantilever is optimized under a reference load $P_0$ and a tenfold larger load $10P_0$, using a compressible Neo-Hookean model and a complementary-work objective. The NTopo baseline is adapted to the same nonlinear energy.

\begin{figure}[!htbp]
\centering\includegraphics[width=\linewidth]{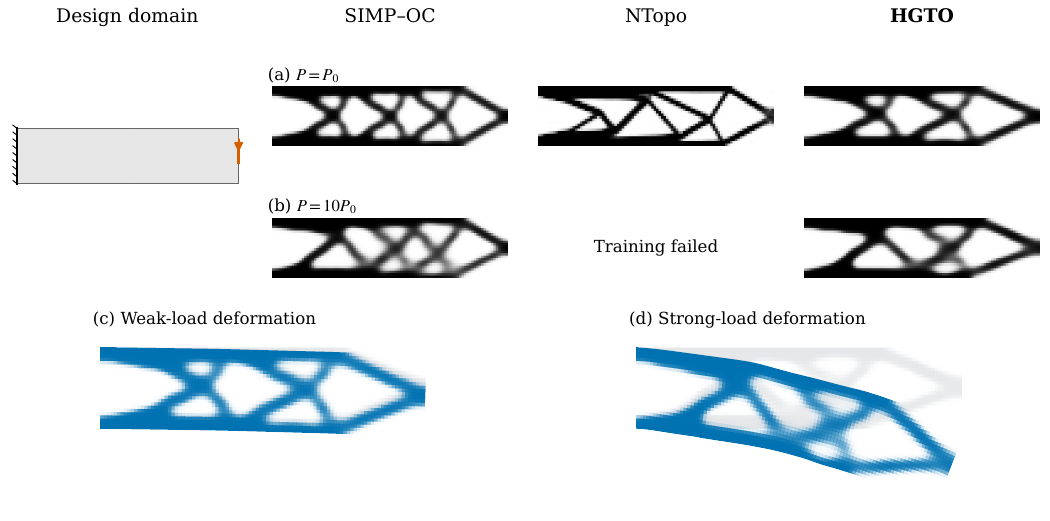}
\caption{Cantilever optimization at (a) $P_0$ and (b) $10P_0$. Panels (c) and (d) show the corresponding HGTO deformations without amplification, with the undeformed designs in gray.}
\label{fig:large_nh}
\end{figure}

At $P_0$, all three methods produce connected cantilever designs. HGTO and SIMP--OC form similar diagonally braced layouts and have closely matched objective values. The dual-field neural baseline finds a more subdivided arrangement with a lower objective value (Fig.~\ref{fig:large_nh}a). HGTO takes 64.1~s, compared with approximately 860~s for either reference method (Table~\ref{tab:nonlinear}).

At $10P_0$, the displacement of the loaded port in the HGTO design reaches approximately 24\% of the span. The deformation in Fig.~\ref{fig:large_nh}d shows substantial rotation of the inclined members and the loaded end. The overall cantilever arrangement is retained, but the member inclinations and junctions change with the load level. Evaluating the evolving geometry allows these changes to enter the density update throughout loading.

HGTO converges in 130.5~s at the stronger load. SIMP--OC reaches a similar objective but stalls after 772.2~s, while the nonlinear NTopo adaptation fails to complete the optimization.

\begin{table}[!htbp]
\centering\small
\caption{Objective $J$ and total optimization time $t$ (s) for finite-deformation problems. NTopo uses the nonlinear adaptation. $\dagger$ Last accepted design before stagnation.}
\label{tab:nonlinear}
\setlength{\tabcolsep}{4pt}
\begin{tabular}{lrrrrrr}
\toprule
 & \multicolumn{2}{c}{SIMP--OC} & \multicolumn{2}{c}{NTopo} & \multicolumn{2}{c}{HGTO} \\
\cmidrule(lr){2-3}\cmidrule(lr){4-5}\cmidrule(lr){6-7}
Case & $J$ & $t$ & $J$ & $t$ & $J$ & $t$ \\
\midrule
Cantilever: $P_0$ & 0.000749 & 854.3 & 0.000660 & 859.7 & 0.000738 & 64.1 \\
Cantilever: $10P_0$ & 0.07640 & 772.2$^{\dagger}$ & \multicolumn{2}{c}{Training failed} & 0.07360 & 130.5 \\
Doubly fixed bridge & 0.11064 & 418.8 & \multicolumn{2}{c}{Training failed} & 0.11086 & 62.0 \\
\bottomrule
\end{tabular}
\end{table}

\subsection{Nonlinear bridge}
\begin{figure}[!htbp]
\centering\includegraphics[width=\linewidth]{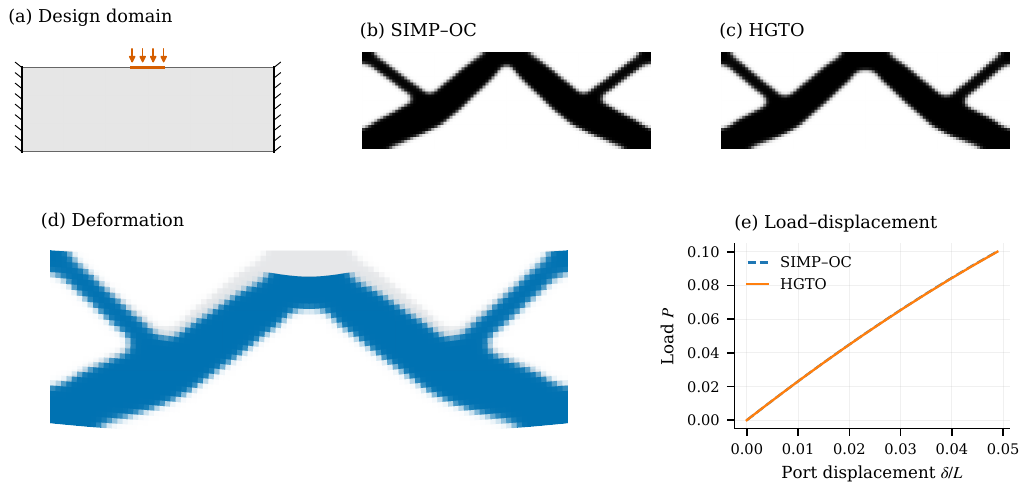}
\caption{Nonlinear bridge example: (a) design domain, (b,c) optimized topologies, (d) HGTO deformation at the final load, and (e) load--displacement curves. Deformation is shown without amplification.}
\label{fig:nh_bridge}
\end{figure}

A doubly fixed bridge subjected to a distributed top load provides a second finite-deformation example. HGTO and SIMP--OC form two principal inclined members beneath the loaded region, with thinner branches extending toward the upper parts of the supports (Fig.~\ref{fig:nh_bridge}). Their final objective values are nearly identical.

The agreement also extends over the loading path. The load--displacement curves overlap closely as the central junction moves downward and the inclined members rotate. Thus, the similar final objectives are accompanied by similar structural responses during loading. HGTO reaches the design in 62.0~s, compared with 418.8~s for SIMP--OC. The nonlinear NTopo adaptation also fails to complete this case.

\subsection{Elastoplastic design}
\label{subsec:plastic}
A perforated connection is optimized using either elastic or elastoplastic response. Both runs use the same material fraction, topology-network configuration, and graph construction, and minimize the peak-load work measure. The elastoplastic model stores and updates material history at the integration points of each hyperedge. The two final designs are then evaluated under a common elastoplastic loading--unloading cycle.

\begin{figure}[!htbp]
\centering\includegraphics[width=\linewidth]{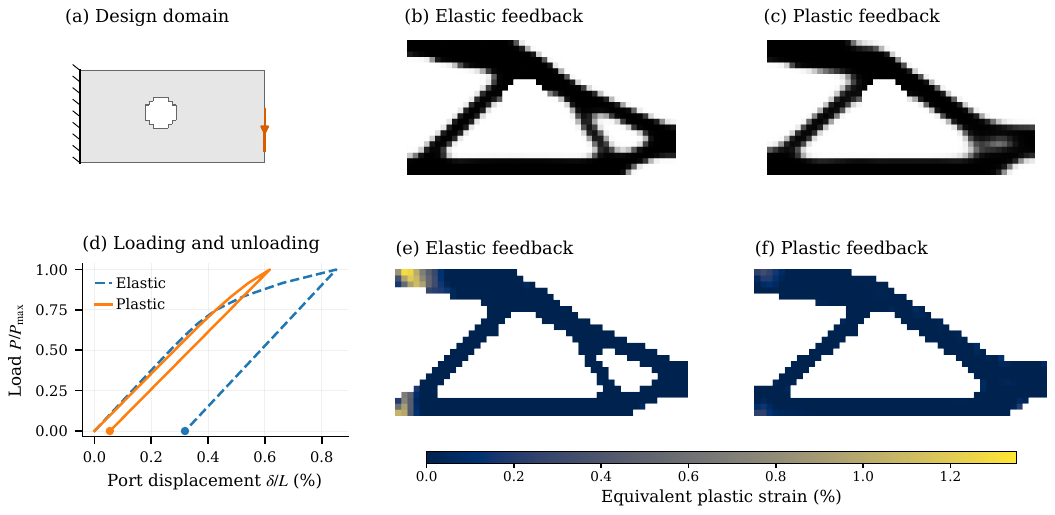}
\caption{Perforated connection optimized with (b) elastic and (c) elastoplastic response. Panel (d) compares both designs under the same elastoplastic loading--unloading cycle; (e,f) show equivalent plastic strain at the peak load.}
\label{fig:plastic}
\end{figure}

The main diagonal and lower member appear in both designs, with local differences near the fixed boundary and the loaded-end junction (Fig.~\ref{fig:plastic}). These differences have a pronounced effect on the response. The elastic-design curve is initially slightly stiffer but bends toward larger displacements as plastic deformation accumulates. The elastoplastic design reaches the peak load with a smaller displacement, reducing the peak-load work by 27.5\%. The ranking based on initial stiffness therefore changes during loading.

The largest differences in the strain fields occur at the fixed ends of the upper and lower members. The elastoplastic design reduces these concentrations, lowering the maximum equivalent plastic strain to approximately one quarter of the elastic-design value. Its plastically active material region is larger in the common evaluation, but the strain is less concentrated. Plastic deformation is therefore distributed over a broader material region at lower intensity.

On unloading, residual displacement at the loaded port decreases from 0.319\% to 0.054\% of the span, a reduction of 83.0\%. This improvement follows from the design obtained under the loading objective. It shows that similar overall topologies can have substantially different permanent deformations, and that incorporating material history can improve the response through changes in member dimensions and connections.

\FloatBarrier
\section{Discussion and conclusions}
\label{sec:discussion}
\label{sec:conclusion}
The numerical studies show that HGTO retains the design flexibility of neural parameterization while achieving compliance comparable to SIMP--OC across the linear benchmarks, irregular domains, and three-dimensional examples. Several planar designs attain this stiffness with fewer secondary members and larger internal openings. The fine-resolution cantilevers preserve these broad arrangements as their boundaries and junctions become more clearly resolved, while the three-dimensional designs form webs, branching supports, and closed hollow sections according to the loading. A compact graph parameterization can therefore accommodate different internal layouts without sacrificing structural performance. Shared network weights coordinate density changes across elements, and mesh adjacency supplies local spatial information. Filtering controls the feature scale, while projection sharpens the solid--void boundary.

For nonlinear design, the quality of mechanical feedback also affects the final structural response. HGTO completes the finite-deformation optimizations as the members rotate and the loaded boundaries move, and incorporates loading history in the elastoplastic connection. In the latter case, the design with slightly lower initial stiffness develops less concentrated plastic strain and substantially less permanent deformation. Its principal members remain similar to those of the elastic-feedback design, but local changes in their dimensions and connections improve the response over the loading--unloading cycle. The benefit of nonlinear feedback is thus evident in the structural response even when the overall topology changes little. The hypergraph operators carry the evolving geometry and material history into the element sensitivities used to train the density field.

HGTO couples material representation and mechanics through a common finite-element structure. Shared element relations connect each density output to its constitutive response, nodal state, and design sensitivity. The mesh thus organizes both the trainable material field and the physical operations that evaluate it. This correspondence is retained as the problem changes: an opening alters the element neighborhoods and nodal connections, whereas a constitutive extension changes the response and internal variables at the physics hyperedges. The irregular and nonlinear examples show that these changes can be accommodated within the same topology-network construction and density--state coupling.

This integration also reduces the cost of neural material design. In the standard planar and three-dimensional comparisons, HGTO completes the optimization in seconds rather than the minutes to approximately one hour required by the representative dual-field neural baseline. Equilibrium is evaluated through prescribed element operations on the nodal states. Shared graph weights keep the trainable design representation compact as the mesh is refined, while cached element data and parallel physical operations support repeated state evaluation. The same 2D network configuration optimizes the 307,200-element cantilever in 195.7~s. HGTO thus combines the flexibility of a learned density field with the computational structure of finite-element mechanics, bringing design quality, efficiency, and adaptability into one differentiable formulation.

The present implementation optimizes each problem separately on a fixed mesh and requires equilibrium evaluation at every density update. Spatial detail and feature size remain linked to the discretization and filter settings, while longer nonlinear loading histories add to the state-evaluation cost. Adaptive meshes and multiscale graph parameterizations could improve the treatment of different spatial scales, and transferring density-network parameters between related problems could reduce repeated optimization. The stresses and internal variables already carried by the physics hyperedges also provide a basis for extending the formulation to local stress constraints and residual-deformation objectives, broadening the structural requirements that can guide neural material design.

\section*{Code availability}
The source code will be made available at \url{https://github.com/Liukz233/HGTO}.

\FloatBarrier
\bibliographystyle{elsarticle-num}
\bibliography{HGTO}

\clearpage
\appendix
\setcounter{section}{0}
\setcounter{figure}{0}
\setcounter{table}{0}
\setcounter{equation}{0}
\renewcommand{\thesection}{S\arabic{section}}
\renewcommand{\thesubsection}{S\arabic{section}.\arabic{subsection}}
\renewcommand{\thefigure}{S\arabic{figure}}
\renewcommand{\thetable}{S\arabic{table}}
\renewcommand{\theequation}{S\arabic{equation}}

\begin{center}
{\Large\bfseries Supplementary Material}

\vspace{0.5em}

{\large HGTO: A Unified Graph-Based Physics-Informed Formulation for Structural Topology Optimization}
\end{center}

\vspace{1em}

\section{Density mapping and optimization settings}
\label{sec:ssettings}
\subsection{Density filtering and projection}
For element centroids $\bm X_e$ and volumes $V_e$, the filter weights and normalized averaging matrix are
\begin{equation}
 w_{ej}=\max\big(0,r_f-\|\bm X_e-\bm X_j\|_2\big),\qquad
 A_{ej}=\frac{V_jw_{ej}}{\sum_kV_kw_{ek}}.
 \label{eq:sfilter}
\end{equation}
The distance weights define the averaging neighborhood, and the volume weights account for unequal element sizes. This is the density filter used for HGTO and SIMP--OC~\cite{bourdin2001filters,andreassen2011efficient}.

For network logits $\bm z$ and a common shift $\lambda_V$, the bounded design variables and physical densities are
\begin{equation}
\begin{aligned}
 x_e&=\operatorname{sigmoid}(z_e-\lambda_V),\qquad \bar{\bm x}=\bm A\bm x,\\
 \rho_e&=\rho_{\min}+(1-\rho_{\min})\mathcal H_\beta(\bar x_e),\\
 \mathcal H_\beta(s)&=
 \frac{\tanh(\beta/2)+\tanh[\beta(s-1/2)]}{2\tanh(\beta/2)}.
\end{aligned}
\label{eq:sprojection}
\end{equation}
The smooth projection sharpens the filtered density as $\beta$ increases~\cite{wang2011projection}. The filter radius remains fixed during projection continuation. The shift is determined by
\begin{equation}
 F_V(\bm z,\lambda_V)=\sum_eV_e\rho_e(\bm z,\lambda_V)-V_fV_\Omega=0.
 \label{eq:svolumeroot}
\end{equation}
Since the density decreases monotonically with $\lambda_V$, this scalar equation is solved by a safeguarded Newton--bisection iteration.

\subsection{Differentiation of the volume constraint}
Let $\bm v=(V_1,\ldots,V_{N_e})^{\mathsf T}$. The density Jacobian with the shift held fixed is
\begin{equation}
 \bm J_0=\operatorname{diag}\!\left[(1-\rho_{\min})\mathcal H_\beta'(\bar x_e)\right]
 \bm A\,\operatorname{diag}\!\left[x_e(1-x_e)\right],
 \label{eq:sJ0}
\end{equation}
where
\begin{equation}
 \mathcal H_\beta'(s)=\frac{\beta\,\operatorname{sech}^2[\beta(s-1/2)]}{2\tanh(\beta/2)}.
\end{equation}
Differentiating Eq.~\eqref{eq:svolumeroot} gives
$\bm v^{\mathsf T}\bm J_0\,d\bm z-\bm v^{\mathsf T}\bm J_0\bm1\,d\lambda_V=0$.
The full density derivative is therefore
\begin{equation}
 \frac{\partial\bm\rho}{\partial\bm z}
 =\bm J_0-\frac{(\bm J_0\bm1)(\bm v^{\mathsf T}\bm J_0)}
 {\bm v^{\mathsf T}\bm J_0\bm1}.
 \label{eq:svolumederivative}
\end{equation}
Its volume-weighted sum is zero, so the linearized density change preserves the prescribed volume. For physical-density sensitivities $\bm g_\rho$, backpropagation evaluates
\begin{equation}
 \bm g_z=\bm J_0^{\mathsf T}\bm g_\rho-
 \frac{\bm g_\rho^{\mathsf T}\bm J_0\bm1}{\bm v^{\mathsf T}\bm J_0\bm1}
 \bm J_0^{\mathsf T}\bm v.
 \label{eq:svolumevjp}
\end{equation}
The products in Eq.~\eqref{eq:svolumevjp} use the sparse filter matrix and pointwise projection derivatives.

\subsection{Network and optimization parameters}
HGTO uses two first-degree Chebyshev layers and fixed Fourier features. The linear 2D studies use 64 frequencies, feature standard deviation 2, and hidden width 64, giving 16,577 trainable parameters. The linear 3D studies use 128 frequencies and width 128, giving 65,921 parameters. Network weights are shared across elements. Each HGTO run starts from a uniform physical-density field with random seed 0. The density floor is 0.001.

For the linear studies, the continuation stages are $(p,\beta)=(1,1),(2,2),(3,4),(3,8)$. The first three stages contain 50 updates each. Adam learning rates decrease from 0.01 to 0.001, with a 200-update reference length for learning-rate decay in the final stage. SIMP--OC uses the same continuation, with move limit $0.2/\max(1,\beta)$.

At the final material parameters, convergence requires a relative objective range below $10^{-3}$ over ten checks and a maximum physical-density change below $5\times10^{-3}$. Both conditions must hold for five consecutive checks, with volume error below $10^{-7}$ and relative equilibrium residual below $10^{-8}$. The total update limit is 1,000. Table~\ref{tab:svolumes} lists the achieved volumes and HGTO update counts.

\begin{table}[!htbp]\centering\small
\caption{Prescribed and achieved volume fractions and HGTO update counts.}
\label{tab:svolumes}
\begin{tabular}{lrrr}\toprule
Case & Target volume & NTopo volume & HGTO updates \\ \midrule
Cantilever $120\times40$ & 0.50 & 0.501115 & 282 \\
Half-MBB & 0.50 & 0.500576 & 280 \\
Inclined load & 0.50 & 0.499636 & 293 \\
Cantilever $240\times80$ & 0.50 & 0.500707 & 257 \\
Cantilever $480\times160$ & 0.50 & 0.500738 & 311 \\
Cantilever $960\times320$ & 0.50 & 0.500734 & 344 \\
L-bracket & 0.40 & 0.400602 & 336 \\
Perforated bracket & 0.40 & 0.400812 & 228 \\
3D cantilever & 0.30 & 0.299229 & 243 \\
Four-foot support & 0.18 & 0.179505 & 257 \\
Torsion member & 0.20 & 0.200417 & 230 \\
\bottomrule\end{tabular}\end{table}

HGTO and SIMP--OC satisfy the volume constraint within $10^{-7}$. All HGTO runs meet the stopping criterion. The inclined-load SIMP--OC run reaches its update limit; the other listed SIMP--OC runs converge.

NTopo follows the original network and training settings~\cite{zehnder2021ntopo}, using hidden widths of 60 in 2D and 180 in 3D, with 200 and 100 outer iterations, respectively. Each configuration uses one initialization, with seed 42 for NTopo. Final linear designs are evaluated with the common material model specified in the main text.

\section{Domains, resolution, and evaluation}
\label{sec:sdomains}
The common linear evaluator uses bilinear quadrilaterals in 2D and trilinear hexahedra in 3D, with full quadrature. Two-dimensional linear cases use plane stress and unit thickness. HGTO mechanics uses FP64 node--element--node operator actions. The optimized physical densities are passed directly to the independent evaluator. All common compliance values use $p=3$, $E_0=1$, $E_{\min}=10^{-6}$, and $\nu=0.3$.

\begin{table}[!htbp]\centering\small
\caption{Linear design domains and HGTO/SIMP--OC filter radii. Grid dimensions count density elements. Radii are expressed in the same length units as the listed domains.}
\label{tab:sdomains}
\begin{tabular}{llll}\toprule
Case & Domain size & Discretization & Radius\\ \midrule
Standard cantilevers / half-MBB & $120\times40$ & $120\times40$ & 3\\
Refined cantilever & $240\times80$ & $240\times80$ & 6\\
Fine cantilever & $480\times160$ & $480\times160$ & 6\\
Finest cantilever & $960\times320$ & $960\times320$ & 6\\
L-bracket & $2\times2$ bounding box & 4,800 quadrilaterals & 0.09\\
Perforated bracket & $3\times2$ bounding box & 3,852 quadrilaterals & 0.12\\
3D cantilever & $48\times16\times16$ & $48\times16\times16$ & 1.6\\
Four-foot support & $24\times32\times24$ & $24\times32\times24$ & 1.6\\
Torsion member & $32\times24\times24$ & $32\times24\times24$ & 1.6\\
\bottomrule\end{tabular}
\end{table}

The standard cantilever has a downward unit force at its right midpoint and a fully fixed left edge. The half-MBB beam restrains horizontal displacement along the left edge and vertical displacement at the lower-right corner, with a downward load at the upper-left corner. The inclined cantilever is loaded at three-quarters of the right-edge height, with the force inclined inward by $25^\circ$ from the downward direction.

For the L-bracket, the upper-right unit square is removed from a $2\times2$ domain. The upper mounting edge is fixed, and the right edge over $0.8\leq y\leq1$ carries a downward unit resultant. The perforated bracket has a central circular clearance of radius 0.5 and a nonuniform conforming quadrilateral mesh. Its left edge is fixed, and the right edge over $0.9\leq y\leq1.1$ carries a downward unit resultant. Removed regions contribute neither material volume nor mechanical energy. The same excluded regions and loaded boundaries are used for the neural baseline.

In 3D, the cantilever has a fixed left face and a $2\times2$ central loading patch on the right. The four-foot support fixes four separate $3\times3$ patches on its lower face and applies a downward unit resultant over a $4\times4$ top pad. Torsion is applied by opposed tractions on two $4\times3$ end patches, with zero net force and a unit moment about the longitudinal axis. All densities in these three domains are design variables.

\begin{table}[!htbp]\centering\small
\caption{Compliance and total computation time for the cantilever resolution study.}
\label{tab:sresolution}
\begin{tabular}{lrrrrrr}\toprule
& \multicolumn{2}{c}{SIMP--OC} & \multicolumn{2}{c}{NTopo} & \multicolumn{2}{c}{HGTO} \\
\cmidrule(lr){2-3}\cmidrule(lr){4-5}\cmidrule(lr){6-7}
Resolution & $C$ & $t$ (s) & $C$ & $t$ (s) & $C$ & $t$ (s) \\ \midrule
$120\times40$ & 175.0823 & 9.02 & 177.9284 & 862.24 & 176.4933 & 8.27 \\
$240\times80$ & 175.7651 & 30.87 & 178.0319 & 862.08 & 176.8524 & 11.72 \\
$480\times160$ & 171.5841 & 197.52 & 178.4512 & 863.68 & 172.4415 & 47.35 \\
$960\times320$ & 168.5552 & 1880.10 & 178.9222 & 867.61 & 170.9963 & 195.75 \\
\bottomrule\end{tabular}\end{table}

The cantilever grids share a 3:1 aspect ratio. The filter radius divided by the span is 0.025 on the two coarser grids, then 0.0125 and 0.00625 on the two finer grids (Table~\ref{tab:sdomains}). For the resolution study, the NTopo results use one density network trained with $150\times50$ samples per batch and evaluated on each listed grid; its reported time includes training and evaluation.

\section{Finite-deformation formulation}
\label{sec:snonlinear}
\subsection{Kinematics and material interpolation}
The finite-deformation examples use a total-Lagrangian description. For reference coordinates $\bm X$ and nodal displacements $\bm u_a$, the deformation gradient at an integration point of element $e$ is
\begin{equation}
 \bm F=\bm I+\bm G,\qquad
 \bm G=\nabla_{\!X}\bm u=\sum_{a\in e}\bm u_a\otimes\nabla_{\!X}N_a .
 \label{eq:sF}
\end{equation}
Here $N_a$ are the element shape functions. The solid material is described by the two-dimensional compressible Neo-Hookean energy
\begin{equation}
 \psi_{\mathrm{NH}}(\bm F)
 =\frac{\mu_0}{2}\left(\bm F:\bm F-2-2\ln j\right)
  +\frac{\lambda_0}{2}(\ln j)^2,\qquad j=\det\bm F>0,
 \label{eq:snhenergy}
\end{equation}
with $\mu_0=E_0/[2(1+\nu)]$ and $\lambda_0=E_0\nu/(1-\nu^2)$. The energy is evaluated with two-dimensional kinematics and has a plane-stress elastic response in the small-strain limit.

Low-density regions are treated with the energy interpolation of Wang et al.~\cite{wang2014finite}. Define the stiffness scale and kinematic switch as
\begin{align}
 s(\rho)&=\frac{E_{\min}+(E_0-E_{\min})\rho^p}{E_0},
 \label{eq:sscale}\\
 \gamma(\rho)&=\frac{\tanh(\beta_0\eta_0)+\tanh[\beta_0(\rho^p-\eta_0)]}
 {\tanh(\beta_0\eta_0)+\tanh[\beta_0(1-\eta_0)]},
 \label{eq:sswitch}
\end{align}
where $\beta_0=500$ and $\eta_0=0.01$. The small-strain companion energy is
\begin{equation}
 \psi_{\mathrm L}(\bm G)=\mu_0\,\operatorname{sym}\bm G:\operatorname{sym}\bm G
 +\frac{\lambda_0}{2}(\operatorname{tr}\bm G)^2.
\end{equation}
The energy used in the density-dependent mechanical calculation is
\begin{equation}
 \psi_\rho(\bm G)=s(\rho)
 \left[\psi_{\mathrm{NH}}(\bm F_\gamma)
       -\psi_{\mathrm L}(\gamma\bm G)+\psi_{\mathrm L}(\bm G)\right],
 \qquad \bm F_\gamma=\bm I+\gamma\bm G.
 \label{eq:sinterpolation}
\end{equation}
For $\gamma=1$, this expression recovers the Neo-Hookean energy scaled by $s$. For $\gamma=0$, it reduces to the scaled linear energy. The switch therefore confines the finite-strain response to material regions while assigning the linear response to low-stiffness regions. Both the HGTO and baseline nonlinear calculations use this interpolation.

\subsection{Physical operators and incremental equilibrium}
Differentiating Eq.~\eqref{eq:snhenergy} gives the first Piola stress and material tangent of the solid,
\begin{align}
 \bm P_{\mathrm{NH}}(\bm F)
 &=\mu_0(\bm F-\bm F^{-\mathsf T})+\lambda_0\ln j\,\bm F^{-\mathsf T},
 \label{eq:sPnh}\\
 (\mathbb A_{\mathrm{NH}})_{iJkL}
 &=\mu_0\delta_{ik}\delta_{JL}
 +\lambda_0 F^{-\mathsf T}_{iJ}F^{-\mathsf T}_{kL}
 +(\mu_0-\lambda_0\ln j)F^{-\mathsf T}_{iL}F^{-\mathsf T}_{kJ}.
 \label{eq:sAnh}
\end{align}
Writing $\bm Q_{\mathrm L}(\bm G)=2\mu_0\operatorname{sym}\bm G+
\lambda_0\operatorname{tr}(\bm G)\bm I$ and
$\mathbb A_{\mathrm L}=\partial\bm Q_{\mathrm L}/\partial\bm G$, the stress and tangent associated with Eq.~\eqref{eq:sinterpolation} are
\begin{align}
 \bm P_\rho&=s\left[\gamma\bm P_{\mathrm{NH}}(\bm F_\gamma)
       -\gamma\bm Q_{\mathrm L}(\gamma\bm G)+\bm Q_{\mathrm L}(\bm G)\right],
 \label{eq:sPrho}\\
 \mathbb A_\rho&=s\left[\gamma^2\mathbb A_{\mathrm{NH}}(\bm F_\gamma)
                         +(1-\gamma^2)\mathbb A_{\mathrm L}\right].
 \label{eq:sArho}
\end{align}
These are prescribed operations on each physics hyperedge. For quadrature weight $\omega_{eg}$, including the reference Jacobian determinant and thickness, the element force and tangent are
\begin{align}
 (\bm f_e)_{ai}&=\sum_g (P_\rho)_{iJ}\,N_{a,J}\,\omega_{eg},
 \label{eq:seforce}\\
 (\bm K_e)_{ai,bk}&=\sum_g N_{a,J}(\mathbb A_\rho)_{iJkL}N_{b,L}\,\omega_{eg}.
 \label{eq:setangent}
\end{align}
Repeated spatial indices are summed. The incidence-based gather and scatter operations assemble these quantities as
\begin{equation}
 \bm f_{\mathrm{int}}=\sum_e\bm P_e^{\mathsf T}\bm f_e,\qquad
 \bm K_{\mathrm T}=\sum_e\bm P_e^{\mathsf T}\bm K_e\bm P_e,
\end{equation}
where $\bm P_e$ extracts element degrees of freedom from the global state. The reference shape-function gradients and incidence are reused as the density and displacement fields evolve.

Loading is applied incrementally, with $\bm f_n=\ell_n\bm f_{\max}$ and $0<\ell_n\leq1$. At each increment, equilibrium on the free degrees of freedom is obtained from
\begin{equation}
 \bm R_n(\bm u_n,\bm\rho)=\bm f_{\mathrm{int}}(\bm u_n,\bm\rho)-\bm f_n=\bm0.
 \label{eq:snhresidual}
\end{equation}
A Newton step solves $\bm K_{\mathrm T}\Delta\bm u=-\bm R_n$, followed by
$\bm u\leftarrow\bm u+\alpha\Delta\bm u$. Backtracking controls the decrease in total potential energy and rejects trial states with nonpositive $\det\bm F_\gamma$. If the Newton direction is not a descent direction, it is replaced by the negative residual rescaled to the same norm before backtracking. Prescribed displacements are imposed directly. The converged state of the preceding load increment initializes the next increment.

\subsection{Design objective and density derivative}
The finite-deformation objective is twice the complementary work at the final load,
\begin{equation}
 J(\bm\rho)=2\left[\bm f_{\max}^{\mathsf T}\bm u_N-U(\bm u_N,\bm\rho)\right],
 \qquad U=\sum_{e,g}\psi_{\rho_e}(\bm G_{eg})\,\omega_{eg}.
 \label{eq:snhobjective}
\end{equation}
For linear elasticity, $U=\tfrac12\bm f_{\max}^{\mathsf T}\bm u_N$ at equilibrium, so this objective reduces to compliance. At a converged nonlinear state, $\partial U/\partial\bm u_N=\bm f_{\max}$. The terms involving $d\bm u_N/d\rho_e$ therefore cancel, giving
\begin{equation}
 \frac{dJ}{d\rho_e}=-2\frac{\partial U}{\partial\rho_e}
 =-2\sum_g\left.\frac{\partial\psi_{\rho_e}}{\partial\rho_e}\right|_{\bm G_{eg}}\omega_{eg}.
 \label{eq:snhgradient}
\end{equation}
The derivative includes both the stiffness interpolation and the kinematic switch. Let
$B=\psi_{\mathrm{NH}}(\bm F_\gamma)-\psi_{\mathrm L}(\gamma\bm G)+\psi_{\mathrm L}(\bm G)$. Then
\begin{equation}
 \left.\frac{\partial\psi_\rho}{\partial\rho}\right|_{\bm G}
 =s'B+s\gamma'\left[\bm P_{\mathrm{NH}}(\bm F_\gamma)
                         -\bm Q_{\mathrm L}(\gamma\bm G)\right]:\bm G,
 \label{eq:snhlocalgradient}
\end{equation}
where
\begin{equation}
 s'=\frac{p\rho^{p-1}(E_0-E_{\min})}{E_0},\qquad
 \gamma'=\frac{\beta_0p\rho^{p-1}\operatorname{sech}^2[\beta_0(\rho^p-\eta_0)]}
 {\tanh(\beta_0\eta_0)+\tanh[\beta_0(1-\eta_0)]}.
\end{equation}
The element sensitivity is propagated through the constrained density map in Section~\ref{sec:ssettings} and the topology network.

\subsection{Problem and optimization settings}
The solid parameters are $E_0=1$, $\nu=0.3$, and $E_{\min}=10^{-6}$. HGTO uses 64 Fourier frequencies, feature standard deviation 2, and hidden width 64. Adam learning rates decrease from 0.01 to 0.001 over a 180-update reference schedule, followed by exponential decay with half-life 20 and floor $10^{-5}$. Gradient clipping is 0.5. Penalization reaches $p=3$ by update 125 and projection reaches $\beta=8$ by update 144. The density floor is 0.001 and the physical filter radius is 1. HGTO and SIMP--OC use twelve load increments, a final relative equilibrium tolerance of $10^{-8}$, and the joint physical stopping rule in Section~\ref{sec:ssettings}.

The cantilever domain is $24\times6$, discretized by $96\times24$ cells. The left edge is fixed. A downward resultant of 0.00125 or 0.0125 is applied to the right-edge port over $2.25\leq y\leq3.75$. The bridge domain is $24\times8$, discretized by $96\times32$ cells, with both side edges fixed. A downward resultant of 0.1 acts over a three-unit segment centered on the upper edge. The bridge averages HGTO logits with their spanwise reflection while retaining the full state mesh.

Nonlinear SIMP--OC uses up to ten step halvings with volume restoration to keep relative objective increases below $10^{-8}$ at unchanged continuation parameters. The strong cantilever terminates with stagnation when no candidate step is accepted, retaining the last accepted density.

The nonlinear NTopo comparison adapts the original training procedure to the finite-deformation model above. The weak-load cantilever completes its 200-iteration budget at volume fraction 0.450644. The strong-load cantilever and bridge terminate during state training after 27.84~s and 13.56~s, respectively, when the deformation determinant becomes nonpositive.

\section{Elastoplastic formulation}
\label{sec:splastic}
\subsection{Plane-strain model and local return mapping}
The perforated connection uses small-strain associative $J_2$ plasticity with linear isotropic hardening~\cite{simo1998computational}. Its displacement field is planar, with $\varepsilon_{zz}=\varepsilon_{xz}=\varepsilon_{yz}=0$. The in-plane engineering strains $[\varepsilon_{xx},\varepsilon_{yy},\gamma_{xy}]$ are embedded in a three-dimensional symmetric tensor using $\varepsilon_{xy}=\gamma_{xy}/2$. The out-of-plane stress is retained in the yield calculation.

At each element density, the same SIMP scale from Eq.~\eqref{eq:sscale} multiplies Young's modulus, initial yield stress, and hardening modulus:
\begin{equation}
 E_e=s(\rho_e)E_0,\qquad
 \sigma_{y,e}=s(\rho_e)\sigma_{y0},\qquad
 H_e=s(\rho_e)H_0.
 \label{eq:sj2scaling}
\end{equation}
Poisson's ratio is constant. This interpolation preserves the yield strain while scaling the local stress response with density. Define $\mu_e=E_e/[2(1+\nu)]$, $\kappa_e=E_e/[3(1-2\nu)]$, and $\lambda_e=\kappa_e-2\mu_e/3$.

The internal variables at an integration point are plastic strain $\bm\varepsilon^p$ and accumulated equivalent plastic strain $a$. Starting from the committed state of load increment $n-1$, the elastic trial stress is
\begin{equation}
 \bm\sigma^{\mathrm{tr}}_n
 =\lambda_e\operatorname{tr}(\bm\varepsilon_n-\bm\varepsilon^p_{n-1})\bm I
 +2\mu_e(\bm\varepsilon_n-\bm\varepsilon^p_{n-1}).
\end{equation}
With $\bm s^{\mathrm{tr}}=\operatorname{dev}\bm\sigma^{\mathrm{tr}}$, $t=\|\bm s^{\mathrm{tr}}\|$, and $\bm n=\bm s^{\mathrm{tr}}/t$ for $t>0$, the trial yield function is
\begin{equation}
 f^{\mathrm{tr}}=t-\sqrt{\frac23}\left(\sigma_{y,e}+H_e a_{n-1}\right).
 \label{eq:sj2yield}
\end{equation}
\Needspace{7\baselineskip}
For $f^{\mathrm{tr}}\leq0$, the step is elastic and the internal variables are unchanged. Otherwise, the radial return gives
\begin{align}
 \Delta\gamma&=\frac{\max(f^{\mathrm{tr}},0)}{2\mu_e+\tfrac23 H_e},
 \label{eq:sreturn1}\\
 \bm\sigma_n&=\bm\sigma^{\mathrm{tr}}_n-2\mu_e\Delta\gamma\bm n,\\
 \bm\varepsilon^p_n&=\bm\varepsilon^p_{n-1}+\Delta\gamma\bm n,
 \qquad a_n=a_{n-1}+\sqrt{\frac23}\Delta\gamma .
 \label{eq:sreturn3}
\end{align}
The flow direction is set to zero when $t=0$. Each integration point stores its own internal variables on the corresponding physics hyperedge.

\subsection{Consistent tangent and equilibrium over a load history}
Let $\mathbb I_{\mathrm{sym}}$ denote the symmetric fourth-order identity and
$\mathbb I_{\mathrm{dev}}=\mathbb I_{\mathrm{sym}}-\tfrac13\bm I\otimes\bm I$. On the plastic branch, define
\begin{equation}
 \vartheta=1-\frac{2\mu_e\Delta\gamma}{t},\qquad
 \bar\vartheta=\frac{1}{1+H_e/(3\mu_e)}-(1-\vartheta).
\end{equation}
The algorithmic tangent of the return map is
\begin{equation}
 \mathbb C_{\mathrm{alg}}
 =\kappa_e\bm I\otimes\bm I
  +2\mu_e\vartheta\mathbb I_{\mathrm{dev}}
  -2\mu_e\bar\vartheta\bm n\otimes\bm n .
 \label{eq:sj2tangent}
\end{equation}
On the elastic branch it reduces to
$\mathbb C_{\mathrm{el}}=\kappa_e\bm I\otimes\bm I+2\mu_e\mathbb I_{\mathrm{dev}}$.
The in-plane block, with engineering-shear conventions, gives the matrix $\bm D_{eg}^{\mathrm{alg}}$ used in the element tangent.

Collect the integration-point histories into $\bm q_n=(\bm\varepsilon^p_n,a_n)$ and denote the return map by $\bm G_n$. At each load increment,
\begin{align}
 \bm R_n(\bm u_n,\bm q_{n-1},\bm\rho)
 &=\sum_e\bm P_e^{\mathsf T}\sum_g\bm B_{eg}^{\mathsf T}
                  \bm\sigma_{eg,n}^{\mathrm V}\,\omega_{eg}-\bm f_n=\bm0,
 \label{eq:sj2res}\\
 \bm q_n&=\bm G_n(\bm u_n,\bm q_{n-1},\bm\rho),
 \label{eq:sj2history}\\
 \bm K_n=\frac{\partial\bm R_n}{\partial\bm u_n}
 &=\sum_e\bm P_e^{\mathsf T}\sum_g\bm B_{eg}^{\mathsf T}
                   \bm D_{eg}^{\mathrm{alg}}\bm B_{eg}\,\omega_{eg}\bm P_e.
 \label{eq:sj2K}
\end{align}
Here $\bm\sigma^{\mathrm V}=[\sigma_{xx},\sigma_{yy},\sigma_{xy}]^{\mathsf T}$, and $\bm B$ maps element displacements to engineering strains. Nodal displacements are updated by Newton iterations using Eq.~\eqref{eq:sj2K}. During those iterations, each local return is recomputed from $\bm q_{n-1}$. The new history is committed only after equilibrium is reached. A new density field is evaluated from the initially stress-free state, with $\bm q_0=\bm0$, over the full prescribed loading sequence.

\subsection{Peak-load objective and history-dependent sensitivity}
The design objective is the peak-load work measure
\begin{equation}
 W_{\mathrm{peak}}(\bm\rho)=\bm f_N^{\mathsf T}\bm u_N,
 \label{eq:sj2objective}
\end{equation}
where $N$ is the final loading increment. This objective measures displacement under the prescribed peak force. Earlier load increments enter through the material history in Eqs.~\eqref{eq:sj2res} and \eqref{eq:sj2history}.

The discrete adjoint is obtained from the augmented functional
\begin{equation}
 \mathcal A=W_{\mathrm{peak}}
 +\sum_{n=1}^{N}\bm\lambda_n^{\mathsf T}\bm R_n
 +\sum_{n=1}^{N}\bm\mu_n^{\mathsf T}(\bm q_n-\bm G_n).
\end{equation}
\Needspace{7\baselineskip}
All equilibrium derivatives below are restricted to free degrees of freedom, and the initial internal state is fixed. Starting from $\bm\mu_N=\bm0$, stationarity with respect to $\bm u_n$ and $\bm q_{n-1}$ yields the backward recursion
\begin{align}
 \bm K_n^{\mathsf T}\bm\lambda_n
 &=-\delta_{nN}\bm f_N+
       \left(\frac{\partial\bm G_n}{\partial\bm u_n}\right)^{\mathsf T}\bm\mu_n,
 \label{eq:sj2adj1}\\
 \bm\mu_{n-1}
 &=-\left(\frac{\partial\bm R_n}{\partial\bm q_{n-1}}\right)^{\mathsf T}\bm\lambda_n
   +\left(\frac{\partial\bm G_n}{\partial\bm q_{n-1}}\right)^{\mathsf T}\bm\mu_n,
 \qquad n>1 .
 \label{eq:sj2adj2}
\end{align}
The density gradient is then
\begin{equation}
 \frac{dW_{\mathrm{peak}}}{d\bm\rho}
 =\sum_{n=1}^{N}\left[
  \left(\frac{\partial\bm R_n}{\partial\bm\rho}\right)^{\mathsf T}\bm\lambda_n
 -\left(\frac{\partial\bm G_n}{\partial\bm\rho}\right)^{\mathsf T}\bm\mu_n\right].
 \label{eq:sj2grad}
\end{equation}
The partial derivatives hold $\bm u_n$ and $\bm q_{n-1}$ fixed and include the density scaling of all three parameters in Eq.~\eqref{eq:sj2scaling}. Local vector--Jacobian products are obtained by differentiating the return map once at each converged load step. Residual derivatives include quadrature weights, whereas derivatives of the pointwise history update do not. The global adjoint solves use the converged consistent tangents. This procedure differentiates the loading history without storing the global Newton iterations. Equation~\eqref{eq:sj2grad} is subsequently propagated through the physical-density map and the topology network.

\subsection{Design and loading--unloading comparison}
The connection has $E_0=1$, $\nu=0.3$, $\sigma_{y0}=0.0015$, $H_0=0.03E_0$, and $E_{\min}=10^{-6}$. The $24\times12$ domain uses a $48\times24$ background grid with 1,100 retained cells after removing the hole centered at $(10.5,6.5)$ with radius 2. The left edge is fixed. A downward resultant of 0.0015 acts on the right edge over $1.5\leq y\leq4.5$. The material fraction is 0.40.

Both designs use 32 fixed Fourier frequencies, feature standard deviation 1, hidden width 64, and uniform initialization. The Adam schedule decreases from 0.003 to 0.0003 over a 180-update reference length, followed by a half-life of 40 and floor $10^{-5}$. The joint stopping criterion and 1,000-update cap are the same as in Section~\ref{sec:ssettings}. The plastic design follows twelve loading increments. The elastic control uses the same elastic constants with yielding suppressed and evaluates the peak load in one step. Both minimize Eq.~\eqref{eq:sj2objective}. The plastic equilibrium tolerance is $10^{-9}$ and the tangent-solve tolerance is $10^{-10}$.

After optimization, both designs undergo twelve loading and twelve unloading increments with the elastoplastic model. This comparison evaluates peak and residual port displacements and equivalent plastic strain under the same loading cycle (Table~\ref{tab:splastic}). The elastic-feedback design has a nominal elastic work of $1.9161\times10^{-4}$, increasing to $3.0642\times10^{-4}$ under elastoplastic evaluation. For the plastic-feedback design, the corresponding values are $2.0280\times10^{-4}$ and $2.2229\times10^{-4}$.

\begin{table}[!htbp]\centering\small
\caption{Elastoplastic response of the two optimized designs under the same loading--unloading cycle. Strain measures refer to material regions ($\rho>0.5$).}
\label{tab:splastic}
\begin{tabular}{lrr}\toprule
Quantity & Elastic feedback & Plastic feedback\\ \midrule
Peak-load work ($\times10^{4}$) & 3.0642 & 2.2229 \\
Peak port displacement / span (\%) & 0.8512 & 0.6175 \\
Residual port displacement / span (\%) & 0.3187 & 0.0541 \\
Maximum equivalent plastic strain (\%) & 1.3513 & 0.3398 \\
Plastically active integration points (\%) & 15.63 & 27.82 \\
Maximum material strain over cycle (\%) & 1.8546 & 0.5901 \\
\bottomrule\end{tabular}\end{table}

Plastic-strain images show element averages of $a_N$ at the peak load, using a common color scale over material regions with $\rho>0.5$. The reported maxima are integration-point values. The plastically active fraction counts integration points satisfying $a_N>10^{-8}$ among those with $\rho>0.5$. Residual displacement is the magnitude of the mean vertical displacement at the loaded port after complete unloading. All normalized displacements use the span of 24.

\section{Execution and timing comparison}
\label{sec:sexecution}
The experiments use an NVIDIA RTX 6000 Ada GPU and an AMD Ryzen Threadripper PRO 5995WX CPU. CPU mechanics uses two threads, and timed jobs run serially. HGTO and SIMP--OC timings cover problem construction, solver and network setup, optimization, and final evaluation. NTopo timings additionally include process startup and output. Plotting and offline evaluation of saved trajectories are excluded.

Linear HGTO uses geometric multigrid preconditioning on structured meshes. The nonuniform perforated mesh uses a CPU-built smoothed-aggregation hierarchy with symmetric GPU V-cycles. The 2D geometric hierarchy is reused for up to ten design updates while the stiffness-change ratio remains at most two. Conjugate-gradient iterations use the current stiffness.

The nonlinear GPU backend uses FP64 tangent assembly and cuDSS 0.8 for sparse LU factorization and forward/adjoint solves. SIMP--OC and the CPU HGTO reference use the same two-thread sparse direct mechanics backend. Final nonlinear designs are evaluated independently with CPU mechanics.

\begin{table}[!htbp]\centering\small
\caption{Nonlinear computation times (s). CPU/GPU denotes the HGTO mechanics backend; its density network runs on the GPU in both versions. The strong-load OC run terminates with stagnation.}
\label{tab:sbackend}
\begin{tabular}{lrrrr}\toprule
Case & OC (CPU) & HGTO (CPU) & HGTO (GPU) & CPU/GPU ratio\\ \midrule
Weak cantilever & 854.28 & 286.90 & 64.10 & 4.48 \\
Strong cantilever & 772.24 & 554.13 & 130.45 & 4.25 \\
Doubly fixed bridge & 418.75 & 377.03 & 61.95 & 6.09 \\
\bottomrule\end{tabular}\end{table}

Table~\ref{tab:sbackend} compares the HGTO mechanics backends with the density network on the GPU in both versions. The objective values agree to numerical precision. GPU mechanics reduces total runtime by factors of 4.25--6.09 relative to CPU mechanics.

\section{Optimization trajectories and visualization}
\label{sec:strajectories}
\begin{figure}[!htbp]\centering
\includegraphics[width=\linewidth]{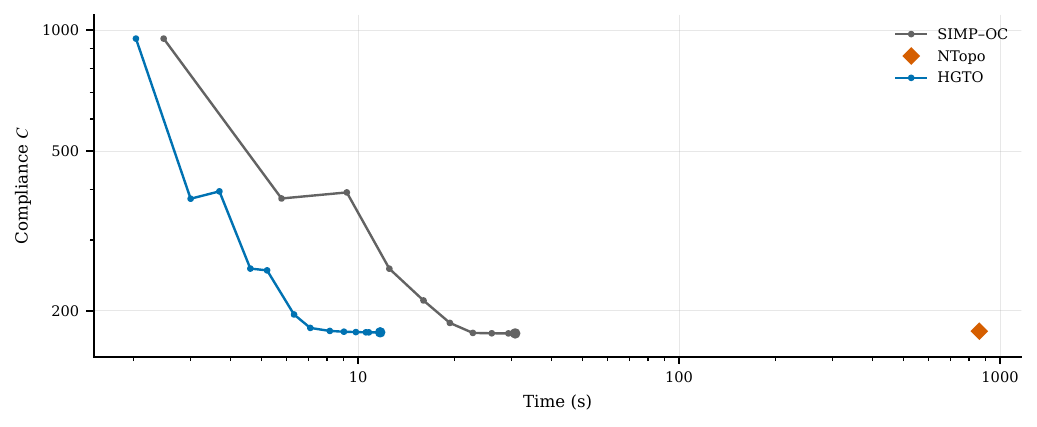}
\caption{Cantilever optimization histories evaluated at the final material parameters. Markers indicate the final designs.}
\label{fig:strajectory}
\end{figure}

Figure~\ref{fig:strajectory} shows the saved $240\times80$ cantilever trajectories. Snapshot compliance is evaluated with the final penalization $p=3$, including snapshots from the continuation stages. Markers show the final designs, with NTopo represented by its endpoint after the prescribed training budget.

Density plots use the optimized physical fields. Three-dimensional surfaces use $\rho=0.5$ and a shared view within each comparison. Nonlinear deformations are shown at their actual scale, with common spatial limits for the weak and strong cantilevers. Load--displacement curves are obtained from the final-design evaluations.

\end{document}